\documentclass[10pt,twocolumn,letterpaper]{article}

\usepackage{iccv}
\usepackage{times}
\usepackage{epsfig}
\usepackage{graphicx}
\usepackage{amsmath}
\usepackage{amssymb}

\usepackage[accsupp]{axessibility}

\usepackage{arydshln}
\usepackage{booktabs}
\usepackage{enumitem}
\usepackage{balance}
\usepackage{combelow}
\usepackage{tabularx}
\usepackage{cite}
\usepackage{comment}

\usepackage{wrapfig}
\usepackage[stable]{footmisc}

\usepackage{multirow}
\usepackage{makecell}
\usepackage{mathtools}
\usepackage{algorithm}
\usepackage{graphicx}
\usepackage{algorithmicx}
\usepackage{eqparbox,array}

\usepackage{xcolor,colortbl}

\usepackage{subcaption}
\usepackage{caption}

\usepackage{setspace}

\usepackage{mwe} % qualitative examples
\usepackage{calc} % qualitative examples
\usepackage{pifont} % checkmark in ablation
\newcommand{\cmark}{\color{blue}{\ding{51}}}
\newcommand{\xmark}{\color{red}{\ding{55}}}

\newcommand{\method}{\mbox{{CAPEAM}}\xspace}
\newcommand{\methodfull}{\mbox{{Context-Aware Planning and Environment-Aware Memory}}\xspace}

\usepackage[pagebackref=true,breaklinks=true,colorlinks,bookmarks=false]{hyperref}

\iccvfinalcopy % *** Uncomment this line for the final submission

\def\iccvPaperID{****} % *** Enter the ICCV Paper ID here

\ificcvfinal\fi

\begin{document}

%%%%%%%%% TITLE
\title{Context-Aware Planning and Environment-Aware Memory \\ for Instruction Following Embodied Agents}

\author{
Byeonghwi Kim\hspace{1.5em}
Jinyeon Kim\hspace{1.5em}
Yuyeong Kim$^{1,}$\thanks{Work done while YK was an intern at Yonsei University. $\dagger$: Corresponding author.}\hspace{1.5em}
Cheolhong Min\hspace{1.5em}
Jonghyun Choi$^\dagger$ \vspace{0.2em}\\
Yonsei University ~~~ $^1$Gwangju Institute of Science and Technology \vspace{0.2em}\\
{\tt\small \{byeonghwikim,jinyeonkim,cheolhong.min,jc\}@yonsei.ac.kr} ~~ {\tt\small uyeongkim@gm.gist.ac.kr}
}
% \author{%
% Anonymous Authors\\
% Anonymous Affiliations \\
% % {\tt\small \{byeonghwikim,jinyeonkim,cheolhong.min,jc\}@yonsei.ac.kr} ~~ {\tt\small uyeongkim@gm.gist.ac.kr}
% ~
% }

\maketitle
\ificcvfinal\thispagestyle{empty}\fi

%%%%%%%%% ABSTRACT
\begin{abstract}
    Accomplishing household tasks requires to plan step-by-step actions considering the consequences of previous actions.
    However, the state-of-the-art embodied agents often make mistakes in navigating the environment and interacting with proper objects due to imperfect learning by imitating experts or algorithmic planners without such knowledge.
    To improve both visual navigation and object interaction, we propose to consider the consequence of taken actions by \textbf{\method} (\methodfull) that incorporates semantic context (\eg, appropriate objects to interact with) in a sequence of actions, and the changed spatial arrangement and states of interacted objects (\eg, location that the object has been moved to) in inferring the subsequent actions.
    We empirically show that the agent with the proposed \method achieves state-of-the-art performance in various metrics using a challenging interactive instruction following benchmark in both \textit{seen} and \textit{unseen} environments by large margins (up to $+10.70\%$ in unseen env.).
    % \method with the templated actions, named ECLAIR, also won the 1$^{st}$ generalist language grounding agents challenge at Embodied AI Workshop in CVPR'23.
\end{abstract}

%%%%%%%%% BODY TEXT
\section{Introduction}
\begin{figure}[t]
    \centering
    \includegraphics[width=.95\columnwidth]{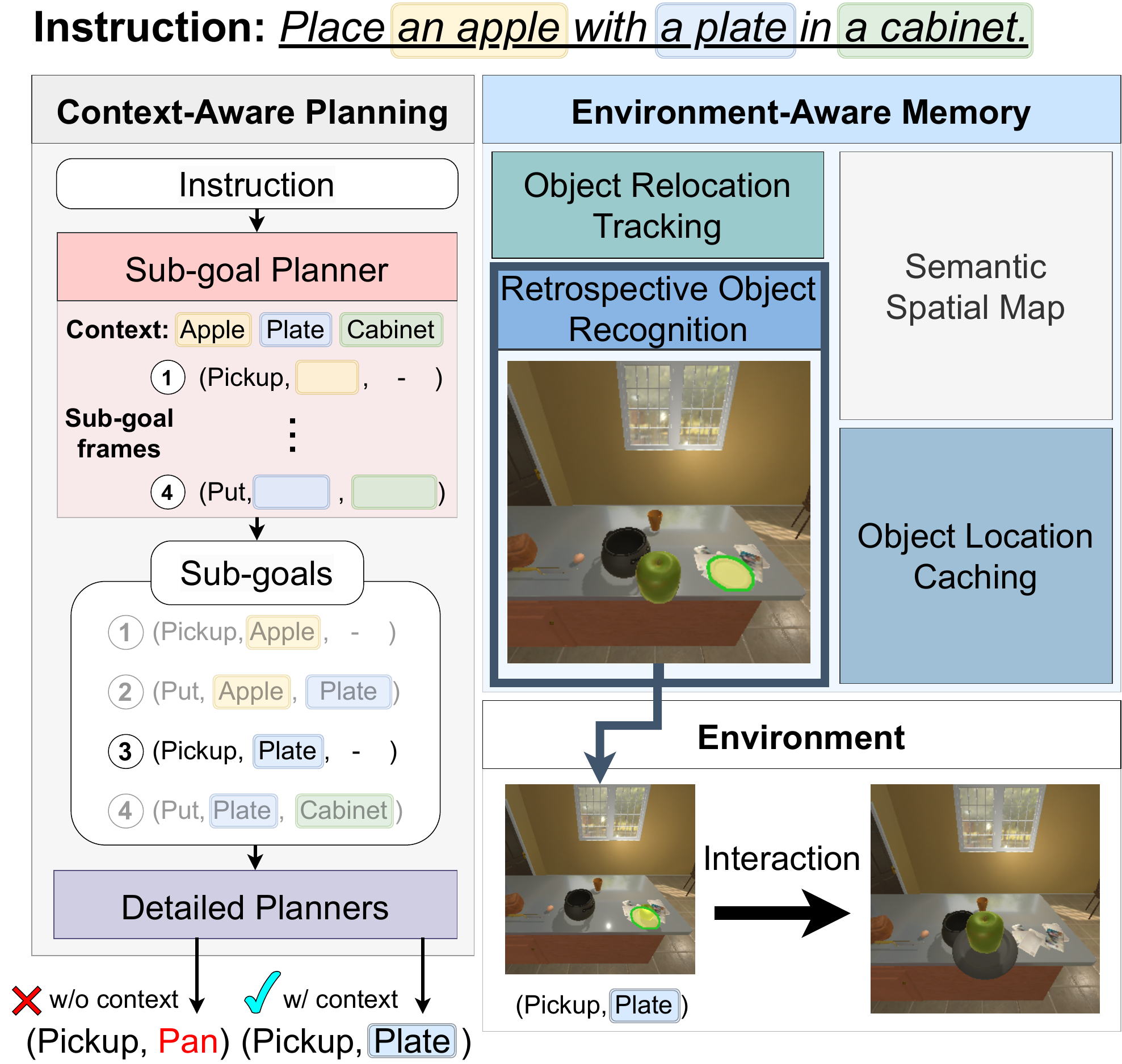}
    % \vspace{-0.5em}
    \caption{
        \textbf{Overview of the proposed `Context-Aware Planning (CAP)' and `Environment-Aware Memory (EAM)'.}
        The CAP incorporates `context' (\ie, task-relevant objects) of the task (denoted by {\cmark} in generating a sequence of sub-goals, compared with the output without the CAP, denoted by {\xmark}).
        The detailed planners then predict a sequence of agent-executable actions for each respective sub-goal.
        The agent keeps the state changes of objects and their masks in the EAM and utilizes them when necessary. 
        Even when the agent may not predict the mask of the plate due to occlusion, it can still interact with the plate thanks to the mask remembered in EAM, leading to successful task completion.}
    \label{fig:teaser}
    \vspace{-1.5em}
\end{figure}

For decades, the research community has been pursuing the goal of building a robotic assistant that can perform everyday tasks through language directives.
Recent advancements in computer vision, natural language processing, and embodied AI have led to the development of several benchmarks aimed at encouraging research on various components of such robotic agents.
These benchmarks include navigation \cite{anderson2018vision,chen2019touchdown,chaplot2017gated,krantz2020navgraph}, object interaction \cite{zhu2017visual,misra2017mapping}, and interactive reasoning \cite{embodiedqa,gordon2018iqa} in visually rich 3D environments \cite{ai2thor,chang2017matterport3d,xia2018gibson}.
However, for realistic assistants to be built, active research in interactive instruction following \cite{misra2017mapping,zhu2017visual,gordon2018iqa,shridhar2020alfred} has been in progress.
This requires agents to navigate, interact with objects, and complete long-horizon tasks by following natural language instructions with egocentric vision.

% Why we need context-aware multi-level planning (CAP)
To accomplish a given task, the agent needs to plan a sequence of actions to interact with specific task-relevant objects.
However, the agent often plans to interact with irrelevant objects to the task.
For instance, for the task ``put an apple slice on the table'', after slicing an apple, the agent might plan to pick up a bread slice, which can lead to the failure of the entire task, mainly due to a lack of contextual memory.
To address this issue, we first propose a novel approach that divides the long-horizon planning process into two distinct phases: 
(1) task-relevant prediction, treated as a \emph{context} prediction, and (2) detailed action planning that considers the contextual memory.
We refer to the term `context' as the objects that the command instructs the agent to manipulate. All actions performed by the agent need to focus on these objects, making them the overarching context for the entire plan's actions.
By prioritizing the prediction of the context, we improve the agent's ability to plan a sequence of actions with less loss of environmental knowledge including objects and their receptacles.
We then combine the generated actions with the context to boost the agent's efficiency in accomplishing long-term objectives by concentrating on interactive objects related to the task.

% Why we need environmental-aware memory (EAM)
In addition, changing the object states poses an additional challenge to the agent's ability to successfully complete tasks that involve object interaction~\cite{zhu2017visual,gordon2018iqa}. % as the states of objects can change during the interaction process \cite{zhu2017visual,gordon2018iqa}.
Failure to track the dynamic object states (\eg, if an object
has been already moved or not) can result in unintended interactions and often lead to task failure.
For example, for the task ``move two apples in the table,'' once the agent moves an apple, the agent might try to move the same apple twice if the agent does not know the apple has already been moved and eventually fails at the task.

To address the additional challenge, we further propose to use an environment-aware memory that stores information about the states of objects, as well as their masks for changed visual appearances mainly due to occlusion.
This approach allows the agent to interact with objects in their proper states over time.
By keeping track of object states and appearances, the agent can ensure interacting with the correct objects and conducting the appropriate actions, ultimately leading to more successful task completion.

% We evaluate our agent in ALFRED and got a new SOTA by large margins ("among published works").
For training and evaluation, we use the widely used challenging benchmark for interactive instruction following \cite{shridhar2020alfred}.
We achieve the state-of-the-art success rates and the goal-condition success rates in \textit{seen} and \textit{unseen} environments by large margins (up to $+10.70\%$ in unseen SR) and rank the first place in the leaderboard at the moment of submission.
Also, \method with the templated approach with minor engineering won the 1$^{st}$ generalist language grounding agents challenge at the Embodied AI Workshop in CVPR 2023.\footnote{See our entry `[EAI23] ECLAIR' in {\url{https://leaderboard.allenai.org/alfred/submissions/public}}}

We summarize our contributions as follows:
\vspace{-0.5em}
\setlength{\leftmargin}{0.8em}
\begin{itemize}
\setlength\itemsep{-0.3em}
    \item We propose context-aware planning that plans a sub-goal sequence with `context' and conducts respective sub-goals with the corresponding detailed planners.
    \item We propose environment-aware memory that stores states in spatial memory and object masks for better navigation and interaction with changed object states.
    \item We achieve a state-of-the-art in a challenging interactive instruction following benchmark \cite{shridhar2020alfred} in all metrics with better generalization to novel environments.
\end{itemize}

\begin{figure*}[t]
    \centering
    \includegraphics[width=\linewidth]{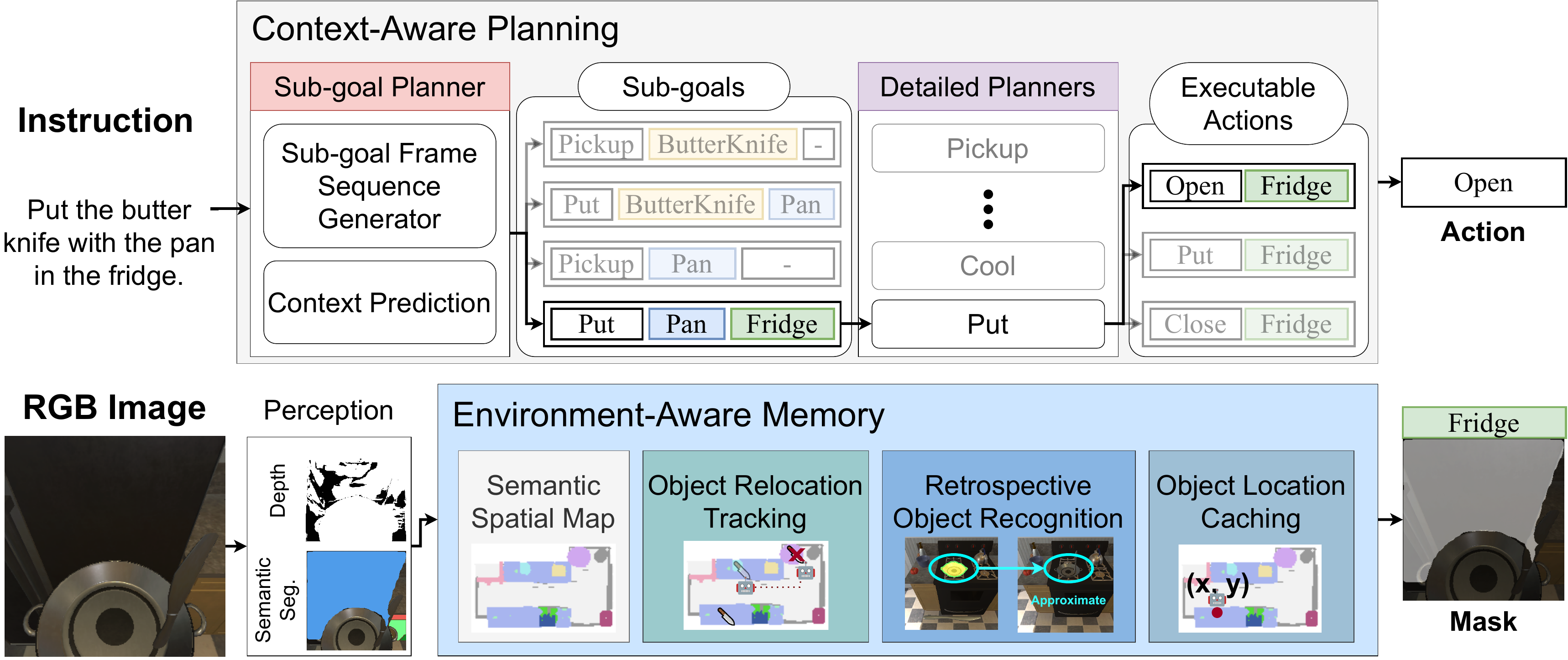}
    % \vspace{-0.5em}
    \caption{
        \textbf{Model Architecture.}
        Our agent consists of (1) `context-aware planning (CAP)' and (2) `environment-aware memory (EAM)'.
        Taking the natural language instructions, the sub-goal planner in the CAP predicts `context' (\ie, task-relevant objects) and generates a sequence of `sub-goal frames' that are sub-goals with a predicted action and placeholders for which object should be used with it.
        Then the objects in the `sub-goal frames' are completed with predicted objects (the context).
        For each planned sub-goal, a corresponding detailed planner generates a sequence of `executable actions.'
        In the EAM, the agent maintains the semantic spatial map by integrating the predicted depths and masks into 3D world-coordinates along with the state changes of objects with their masks to utilize them during task completion.} 
    \label{fig:overview}
    \vspace{-0.5em}
\end{figure*}

\section{Related Work}

% Short Introduction
% Interactive instruction following is a challenging problem that requires vision and language multi-modal understanding and complex reasoning for task completion~\cite{shridhar2020alfred}.
% In recent years, the problem has gained significant attention and we have witnessed great advancement in architectural design or employing neuro-symbolic modules~\cite{singh2021factorizing,pashevich2021episodic,nguyen2021look,suglia2021embodied,zhang2021hierarchical,song2022one,blukis2021persistent,ishikawa2022moment,min2021film,jia2022learning,murray2022following,liu2022planning,inoue2022prompter}.

\paragraph{Action Planning.} %Multi-Level Reasoning.}
Embodied AI tasks require agents to reason at multiple levels of abstraction, taking into account the dynamic nature of the environment, their capabilities, and goal formulation~\cite{embodiedqa,gordon2018iqa,weihs2021visual,ehsani2021manipulathor,padmakumar2022teach}.
Despite the complexity of tasks, many approaches \cite{singh2021factorizing,pashevich2021episodic,nguyen2021look} rely on flat reasoning that directly outputs low-level actions.

Prior arts \cite{gordon2018iqa,das2018neural,corona2020modular,zhang2021hierarchical,blukis2021persistent} attempted to address this issue by splitting the layer of actions into two, with the first layer composed of abstract natural language instructions and the second layer of agent-interpretable low actions.
However, a significant semantic gap exists between natural language instruction and agent-interpretable actions.
Several approaches \cite{landi2019embodied, krantz2020navgraph, pashevich2021episodic} require large amounts of labeled data or trial-and-error learning to bridge this gap.
\cite{suvaansh2023multi} propose a deep model to reduce the semantic gap issue without requiring excessive labeled data or trial-and-error learning.
However, their models often suffer from confusion in predicting the correct objects to interact with.

Meanwhile, the templated approaches~\cite{min2021film, inoue2022prompter} have been explored for data-efficient action planning.
They rely on pre-designed templates for every task type in the dataset and match each task to the corresponding template.
Despite the benefits of efficiency and accuracy, human experts must generate templates whenever a new task type is needed, which is time-consuming and resource-intensive.
Additionally, this may not generate optimal plans, as it is restricted to predefined protocols and may not be suitable for tasks outside of the specified protocol.
% Therefore, the template-based approach has potential drawbacks in dealing with more complex and varied tasks.
Instead, we propose a method that takes full advantage of the deep learning model to decrease the rigidity of the templated approaches.

For effective planning, it would help agents focus on a task context to parse task-relevant objects from natural language instructions~\cite{chen2022weakly,georgakis2022cross}.
\cite{chen2022weakly} learn to localize instruction-relevant objects based on implicit language representation.
\cite{georgakis2022cross} uses the implicit representation of language instructions to guide hallucination of the region outside of the input egocentric map.
However, such implicit representation can be affected by language ambiguity (e.g., a red one, the fruit, and apple for ‘Apple’) as the agents are not supervised which they refer to as the same object.
In contrast, we explicitly predict task-relevant objects (e.g., Apple, Knife, etc.) that can help the agent focus on the context and thus be robust to variation of language instruction.

% \subsection{Planning with Large Language Models}
Aside from the learning-based approaches for planning, large language models (LLMs)~\cite{brohan2023can,huanginner,huang2022language,raman2022planning} have been actively investigated for their efficacy in reasoning.
Despite their significant 
In \cite{huang2022language}, LLMs lack physical grounding due to the absent connection between the physical state of agents and environments (e.g., executable actions, environmental consequences of actions, etc.).
This may result in unreasonable interpretations of instructions.
For instance, for the task of ‘turn on the lamp while holding a tennis racket,’ the LLM may generate a plausible plan: 1) pick up the tennis racket, 2) plug in the code of the lamp, and 3) turn on the lamp.
However, the agent may not support the ‘plug-in’ action, leading to failure at the task.
To address the physical grounding issue in LLMs, efforts have been made including reprompting \cite{raman2022planning}, success detectors \cite{huanginner}, and skill affordance value functions \cite{brohan2023can}, which implies that sub-goal planning with LLMs still requires further investigation.

\vspace{-1em}
\paragraph{Memory for Object Interaction.}
% Put the contents of Remove_1st_Obj to here
Utilizing the semantic spatial map makes the process of searching for objects efficient. 
Previous works \cite{min2021film,inoue2022prompter} record the objects' positional information on the map when the agent finds key objects during exploration and uses it when the current sub-goal matches one of these records. 
However, these data do not contain whether the interaction has already been completed or the object should not be moved again. 
This may cause the agent to re-interact with the object that should not be interacted with. 
To address this issue, we propose environment-aware memory that keeps track of objects' information if it has been interacted with, which can reduce misinteraction with unsuitable objects.

In addition to spatial representation encoding, interaction with the same object in multiple time steps is also challenging as the appearance of an object may change during the interaction (\eg, opened drawer vs. closed drawer) but the agent has to recognize the same object with different appearances.
For multiple interactions with the same object, \cite{singh2021factorizing} proposes instance association in time (IAT) that introduces a memory to store the previous time step's mask for an additional mask-selection criterion based on the geological distance of masks to select the object mask closest to the previous time step's one.
However, if the mask generator fails to recognize the object, the IAT then has no mask candidates for selection and therefore the agent is not able to choose the current object's mask.
In contrast, we propose to memorize the previously interacted masks to reduce the effect of the appearance changes mainly due to occlusion.

\vspace{-1em}
\paragraph{Semantic Spatial Representation.}
The earlier strategies to the problem~\cite{singh2021factorizing,pashevich2021episodic,nguyen2021look} are to directly map visual observations and natural language instructions to a sequence of low-level actions and corresponding masks for object interaction by encoding their history in implicit representation. % \cite{singh2021factorizing,pashevich2021episodic,nguyen2021look}
For instance, \cite{singh2021factorizing} proposes to use separate network branches for action and mask prediction and each branch learns to map visual and language features to a sequence of respective actions and masks. % \cite{singh2021factorizing}
\cite{pashevich2021episodic} proposes a Transformer-based agent that jointly learns to predict actions and masks based on the history of visual observations and language instructions. % \cite{pashevich2021episodic}
While they outperform the baseline~\cite{shridhar2020alfred} with large margins, they still lack the ability to complete tasks in unseen environments as observed in the performance gap from seen environments. % \cite{shridhar2020alfred}

Multiple recent works propose to build an explicit semantic spatial representation such as 2D top-down maps~\cite{min2021film,inoue2022prompter}, 3D voxel maps~\cite{blukis2021persistent,ishikawa2022moment,murray2022following}, or graphs~\cite{liu2022planning}.
Such representation enables the agent to accurately perceive 3D environments and plan actions to navigate to and interact with objects on the representation at the expense of additional supervision.
Inspired by this, we maintain our agent's history in a semantic map for room layouts, objects, \etc.

\section{Approach}
Generalizing a learned embodied AI agent to the unseen environment is one of the key challenges of building a successful embodied AI agent~\cite{blukis2021persistent,min2021film,murray2022following,inoue2022prompter}.
To achieve better generalization, recently proposed successful approaches \cite{blukis2021persistent,ishikawa2022moment,murray2022following} often build an explicit spatial map for the environment and use a hybrid approach of combining learning the environment with well-designed navigation strategies.

Despite being successful, these agents often forget their task contexts and therefore attempt to interact with task-irrelevant objects, eventually leading to task failure.
In addition, they might encounter different objects' states (\eg, appearances, positions, \etc) due to object interaction, which may require tracking the states while completing tasks.

To address this issue, we introduce context-aware planning that plans a sequence of actions based on a context (\ie, task-relevant objects) and environment-aware memory that stores states in a spatial memory and object masks for better navigation and interaction with the objects in various states (\eg, an object containing another object inside) in a hybrid model of learning and crafted navigation algorithms.
Figure \ref{fig:overview} illustrates the architecture of our \method.
We provide details for each component of our method below. % below. %in the subsequent sections.

\subsection{Context-Aware Planning} % Hierarchical Language Understanding (Meta Controller, Manipulators, etc.)
\label{subsection:context_aware_multi_level_planning}
Upon receiving a natural language command, an agent needs to interpret and infer the requirements of the given task (\eg, to fetch the object of interest). 
Once the agent successfully interprets the command, the agent needs to create a plan to achieve the goal.
Similar to \cite{zhang2021hierarchical,blukis2021persistent,suvaansh2023multi}, we propose a novel planning scheme that divides the goal into `sub-goals' and develops each sub-goal into a `detailed action sequence' that the agent can execute.

However, generating a proper sub-goal sequence for the complex task is not trivial as the generated sub-goals often contain irrelevant objects or undesirable actions.
Particularly, if there is no proper error-correcting mechanism in place, which is not trivial either, an incorrect sub-goal leads to a failure of the entire task.
Even if the agent corrects the error, the corrections take extra steps to complete the task, harming the efficiency of the agent. 

For the correct planning with task-relevant objects, we first define `context' as a set of task-relevant objects shared across sub-goals of a given task. 
The proposed `context-aware planning' (CAP) divides planning into two phases; 1) a `sub-goal planner' which generates sub-goals, and 2) a `detailed planner' which is responsible for a sequence of detailed actions and objects for interaction for each sub-goal.
The sub-goal planner further comprises two sub-modules: the context predictor, which predicts three task-relevant objects, and the sub-goal frame sequence generator, which generates a sequence of sub-goals that do not rely on particular objects, referred to as sub-goal frames.

There are three task-relevant objects referred to as the context.
The first object corresponds to the main object to be manipulated. 
The second object pertains to the container that holds the object. 
The last object relates to the target object where the object is to be placed in the task. 
We integrate these predicted task-specific objects into a sequence of sub-goal frames to produce a sub-goal sequence.
This allows our agent to plan a sequence of sub-goals conditioned on the task-relevant objects, which helps the agent remember the context, \ie, objects, during action planning.

\begin{figure}[t]
    \centering
    \includegraphics[width=\columnwidth]{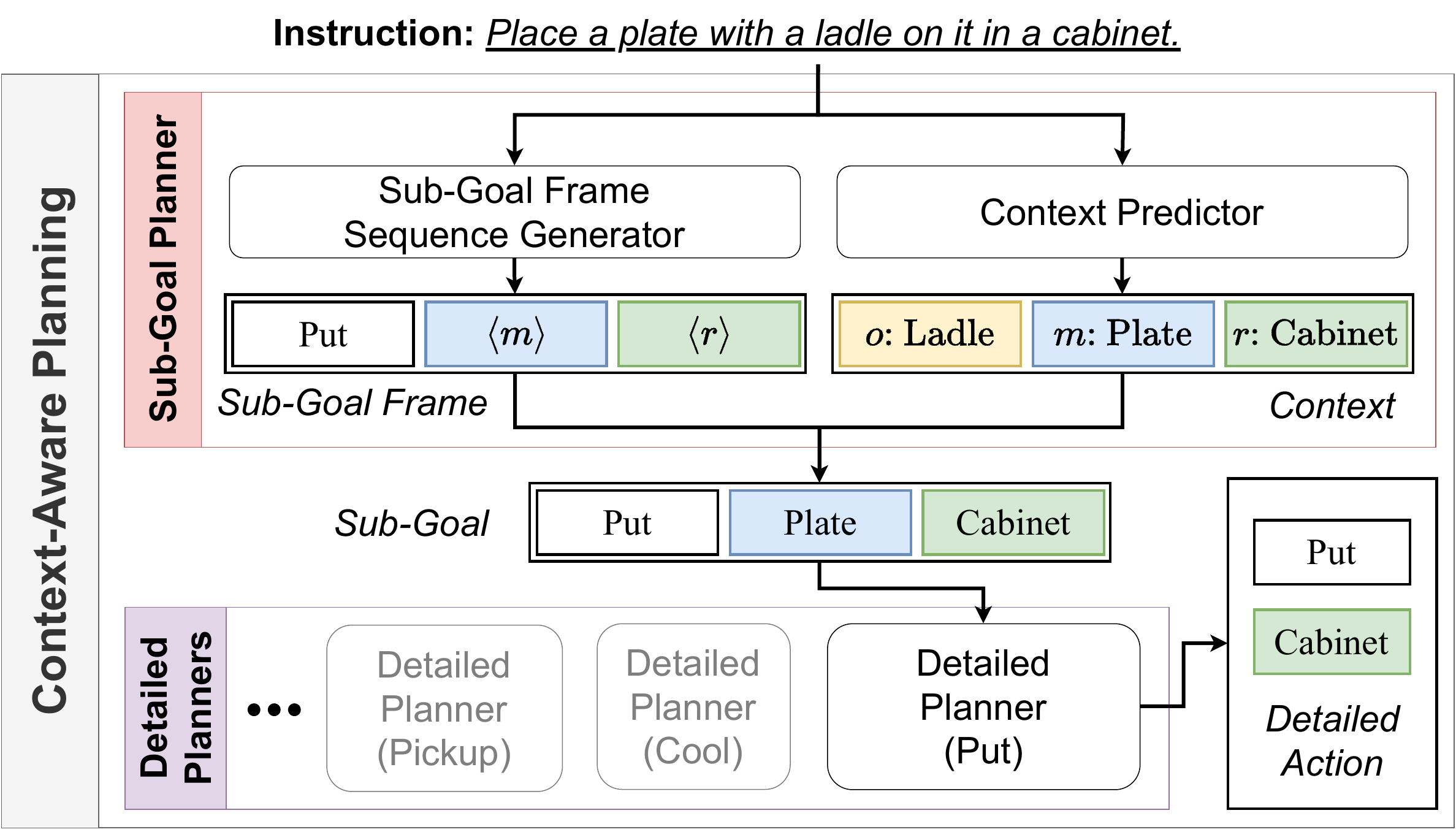}
    \vspace{-1.5em}
    \caption{
        \textbf{Context-Aware Planning (CAP).}
        It consists of a `sub-goal planner' and a set of `detailed planners' for each sub-goal to generate `executable actions.'
        The sub-goal planner first predicts a set of objects related to the task, which we call `Context.'
        Then, the `sub-goal frame sequence generator' in the sub-goal planner generates a sequence of `sub-goal frames.'
        Finally, the `meta-classes' in each sub-goal frame are replaced with the corresponding objects in the context, resulting in the final sub-goal.
        A `detailed planner' translates the sub-goal to executable actions.} % for the agent to execute.}
    \label{fig:cp}
    \vspace{-1em}
\end{figure}

\subsubsection{Sub-Goal Planner} % separate action prediction and argument prediction
Given a language input $l$, the sub-goal planner, $f_{sub}(\cdot)$, generates human-interpretable sub-goals.
We can write the $n^\text{th}$ sub-goal as a triplet of action, a small object, and a receptacle that contains the object to be interacted with as:
\begin{equation}
\begin{split}
    f_{sub}(l) &= \{S_n\}_{n=1}^N,\\
    S_n &= (A_n, O_n, R_n),
\end{split}
\end{equation}
where, $A_n$ denotes a human-interpretable action, such as `clean' or `heat'. 
$O_n$ denotes a small object targeted for manipulation in the execution of $A_n$. 
$R_n$ refers to the location where $O_n$ can be found. 
$N$ refers to the number of sub-goals in a plan.
We have two sub-modules to generate the triplets: the context predictors and the sub-goal frame sequence generator.
The context predictors focus on extracting task-relevant objects from natural language instruction. 
The sub-goal frame generator concentrates on a sequence of interactions that constitute the goal.

We predict task-relevant objects to ensure that all sub-goals in the plan share the same task-relevant object information. By utilizing a `meta-class' to generate sub-goal sequences that identify where these task-relevant objects should be placed. 
This ensures that all sub-goals share the same task-relevant objects. 
We refer to a sub-goal filled with the meta-classes as a `sub-goal frame'. % task-relevant objects to be inserted into.

% Context prediction
\vspace{-0.5em}
\paragraph{Context Prediction.}
\label{subsec:ContextPrediction}
Given a human-described instruction $l$ as an input, we use three context predictors.
$f_{ctxt}^O(\cdot)$ predicts a primary object being targeted in the task, noted as $c_O$.
$f_{ctxt}^M(\cdot)$ predicts a required carrier of the object, noted as $c_M$.
$f_{ctxt}^R(\cdot)$ outputs a destination (\ie, receptacle) that contains the target object, referred to as $c_R$:
\begin{equation}
    f_{ctxt}^O(l) = c_O,~~ f_{ctxt}^M(l) = c_M,~~ f_{ctxt}^R(l) = c_R,
\end{equation}
where $c_O$ is an object assumed to play the main target in a task described as $l$, $c_M$, and $c_R$ are a container and destination of $c_O$, respectively.
For instance, if the goal is to ``place an apple in a mug on a table'', the agent needs to move the `apple' ($c_O$) using the container `mug' ($c_M$) and subsequently place both of them onto the `table' ($c_R$). 
To predict the context, we finetune a pretrained language model~\cite{devlin2018bert}.
% For a fair comparison with prior arts \cite{min2021film,inoue2022prompter}, we use the same pretrained networks in a supervised manner to predict each task-relevant object.

% Action planning
\vspace{-0.5em}
\paragraph{Sub-Goal Frame Sequence Generator.}
After identifying the objects that require manipulation in the task, we determine a series of interactions that will bring objects to the desired goal state.
To do this, we generate a sequence of sub-goals \emph{without} specifying the context (\ie, objects) involved. 
This allows the sub-goals to be later filled with predicted task-relevant objects.
This task is accomplished through the sub-goal frame sequence generator.

% todo: object 대신 meta 이유
We introduce a meta-class, which is mapped to one of the corresponding contexts. 
The use of a meta-class instead of the specific object name allows the sub-goal frame sequence generator to focus more on the object's role (\ie, a main target, a carrier, and a destination) in the task rather than its own name, as illustrated in Figure~\ref{fig:qualitative_cap}.
If a sub-goal frame includes meta-classes ($x_O$, $x_M$, $x_R$), it is later replaced with the `contexts' ($c_O$, $c_M$, $c_R$) from the context predictors.
Note that some contexts may contain `None' indicating that a task does not need them.
For instance, for a task, `put an apple on the table,' the carrier ($x_M$) is not needed and therefore an action sequence is planned based only on the two contexts (\ie, $x_O=\text{Apple}$ and $x_R=\text{Table}$).

The sub-goal frame outputs a sub-goal with a placeholder ($\langle \cdot \rangle$), which is later filled with either an object or a meta-class by the generator.
Formally, the sub-goal frame sequence generator $f_{sf}(\cdot)$ takes as input a natural language instruction and generates a sequence of sub-goal frames:
\begin{equation}
\begin{split}
    f_{sf}(l) &= \{F_n\}_{n=1}^N,\\
    F_n &= (A_n, \langle O \rangle_n, \langle R \rangle_n),\\
    \langle \cdot \rangle &\in E \cup \{x_O, x_M, x_R\},\\
\end{split}
\end{equation}
where $l$ denotes the instruction, $A_n$ denotes an action of $n^{th}$ sub-goal frame, $\langle O \rangle_n$ and $\langle R \rangle_n $ represent place holder of the small object and receptacle to be interacted with, respectively, $E$ refers to a set of all objects in a given environment, and $N$ refers to the number of sub-goal frames.
All meta-class in place holder of sub-goal frames are then replaced with task-relevant objects ($c_O$, $c_M$, $c_R$) from context prediction, resulting in the final output $\{(A_n, O_n, R_n)\}_{n=1}^N$.

\subsubsection{Detailed Planners}
To execute each sub-goal generated by the sub-goal planner $f_{sub}$, an agent should render the agent-executable actions which we refer to as `detailed actions' from the inferred `sub-goal plan', $\{S_n\}_{n=1}^N$.
We define a `detailed planner' $f_{dp}^g(\cdot)$ for a sub-goal action $g$ to translate a sub-goal with $A_n = g$ into a sequence of the detailed actions as:
\begin{equation}
    f_{dp}^g((A_n, O_n, R_n)) = \{(a_t, o_t)\}_{t=1}^{T_n},
\end{equation}
where $a_t$ and $o_t$ are the $t^\text{th}$ detailed action and an object to interact in the given sub-goal.
For instance, if a sub-goal is given as (Pickup, Plate, Cabinet), a favorable output of a detailed planner would be (Open, Cabinet), (Pickup, Plate), and (Close, Cabinet).
We learn the detailed planner using a self-attention LSTM in a supervised manner~\cite{katrompas2022enhancing}.

\begin{figure}[t]
    \centering
    \includegraphics[width=\columnwidth]{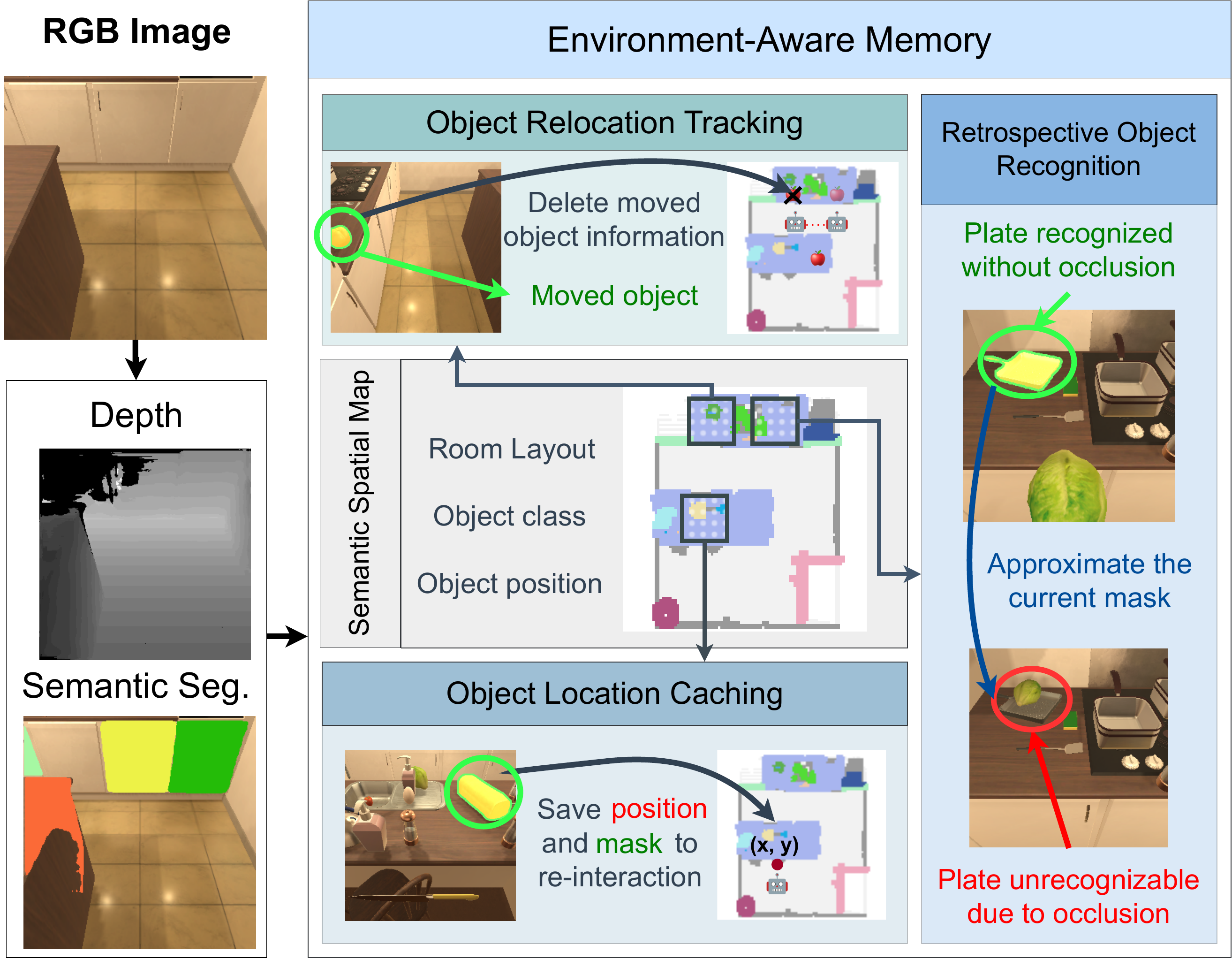}
    \vspace{-1.5em}
    \caption{
        \textbf{Environment-Aware Memory (EAM).}
        The agent updates the semantic spatial map using predicted depths and object masks for scene information.
        `Retrospective Object Recognition' preserves the latest object mask to approximate the current object's mask when mask prediction fails.
        `Object Relocation Tracking' stores the most recent location of each relocated object and discards it as a future navigation target.
        `Object Location Caching' remembers the locations and masks of objects whose states change.
    }
    \label{fig:environment_aware_memory}
    \vspace{-1em}
\end{figure}

\subsection{Environment-Aware Memory}
\label{subsection:environment_aware_memory}

\paragraph{Semantic Spatial Map.}
One of the challenges in the interactive instruction following task is to accurately perceive the 3D environment from the 2D RGB images captured by the agent for better navigation and interaction with objects.
Following \cite{blukis2021persistent,ishikawa2022moment,min2021film,murray2022following,inoue2022prompter}, we build a semantic spatial map from predicted depth maps and object masks by back-projecting them to 3D world coordinates.
In particular, we use depth to maintain the environmental information such as obstacle area, object positions and classes, \etc.
% Inspired by them, we let our \method maintain the observed scene information in the semantic spatial map.

However, without memory, the agent may not be able to keep track of the configurations of objects in an environment.
The absence of such tracking poses an additional challenge for task completion. % as the agent is often required to utilize them.
For instance, if the agent is asked to move multiple apples to a table, the agent should then remember which objects have been moved so far.
Here, we propose to configure memory of past environmental information to remember the past information and predict a proper action sequence during a task. 
The memory helps the agent remember the current state of the environment and make proper decisions to facilitate task completion. 
% By maintaining such memory, the agent can track the progress of the task, remember previous instructions, and reason about the current state of the environment.
% This is particularly important in complex tasks that involve multiple sub-tasks or steps, where the agent needs to remember the context of each step and how it relates to the overall goal.

\vspace{-1em}
\paragraph{Retrospective Object Recognition.}
\label{subsec:Retrospective_Object_Recognition}
While completing tasks, the agent often needs to interact with the same object in multiple time steps.
For example, to move an apple with a plate, the agent has to interact with the plate twice: 1) put the apple in the plate and 2) pick up the plate with the apple.
During multiple object interactions, however, the visual appearance of the object can change due to various reasons such as occlusion. %, changed visual appearances, \etc.
For these reasons, the agent might not be able to recognize the object and would fail at interaction with the intended object, leading to the entire task's failure.

To address changes in the visual appearance of an object during multiple interactions, we propose to retain the latest segmentation masks of objects and use them as the current object's mask if the agent is interacting with the same object but fails to recognize it.
Exploiting the preserved masks allows the agent to keep interacting with the same object even with visual appearance changes during interactions.

\vspace{-1em}
\paragraph{Object Relocation Tracking.}
In another task completion scenario, the agent often faces scenarios where it needs to relocate multiple objects of the same class to the same location (\eg, ``Move two apples on the table.'').
While conducting this task, the agent may lose track of which objects have already been relocated.
Consequently, the agent may attempt to navigate to and relocate objects that have already been relocated, which possibly leads to a repeated sequence of unintended object relocation or a task failure.

To circumvent the issues, we propose to maintain the information about the most recent location of each relocated object in a 2D coordinate and exclude it in the semantic map as a future target for our agent's navigation.
The agent can recognize relocated objects among all detected objects by comparing the locations in the memory and the semantic spatial map.
%detecting all the apples and checking the memory states related with each apple
This module allows the agent to avoid redundant interaction with already relocated objects, possibly leading to task failure, and therefore successfully navigate to and interact with those that have been not relocated.

\vspace{-1em}
\paragraph{Object Location Caching.}
%Preserving information related to past events can help the agent avoid redundant actions and task failures in retracing actions.
%When performing tasks that involve multiple steps, an agent must keep track of previously completed actions and the objects that were manipulated.
% For instance, in the example of "putting an apple slice on the table," an agent needs to cut the apple, leave the knife in another location, and return to the cutting position to retrieve the sliced apple.
During another task completion, an agent may need to revisit object locations that it has previously visited to obtain an object with `changed states'.
Considering the task ``put an apple slice on the table'' as an example, after the agent has sliced the apple, the agent needs to navigate to the sliced object again at the location where the apple was sliced to interact with the object in its changed state.
Without memorizing the locations and masks, however, the agent would need to re-explore the environment to locate the objects again, which could result in inefficient navigation and possible task failure.

To alleviate such an issue, we propose to cache the 2D locations and the segmentation masks in memory for objects whose states change.
By preserving the object locations and masks, the agent can efficiently navigate back to the remembered locations and interact with the remembered object masks when necessary.
This can reduce the need for the agent to explore the environment again, which can lead to more efficient navigation and interaction and possibly reduce the possibility of navigation and interaction failure. % for the target object.

\subsection{Action Policy} % relate this to context-aware planning?
To conduct object interaction for task completion, the agent first needs to reach the target objects in close vicinity \cite{ramrakhya2022habitat,weihs2021visual,padmakumar2022teach}.
Recent approaches \cite{blukis2021persistent,ishikawa2022moment,murray2022following,inoue2022prompter} plan obstacle-free paths using either deterministic algorithms (\eg, A*, FMM \cite{sethian1996fast}, \etc) or learning the path in the discrete navigation space~\cite{tamar2016value}, mostly by the imitation learning~\cite{bain1995framework}.

Since imitation learning requires a large number of expert trajectories for satisfactory performance, deterministic algorithms currently dominate the literature for significantly better performance, implying the amount of data in the current benchmark with imitation learning might be more limited than necessary.
In addition, \cite{min2021film,inoue2022prompter} maintains the obstacle area larger than they actually perceive for safe path planning distant from the obstacles.
Inspired by them, we plan navigation paths with the deterministic approach \cite{sethian1996fast} in the discrete space on the expanded obstacle map.

\section{Experiments}
\label{sec:experiments}

\paragraph{Dataset and Metrics.}
We employ ALFRED~\cite{shridhar2020alfred} as a challenging interactive instruction following benchmark.
There are three splits of environments in ALFRED: `train', `validation', and `test'.
The validation and test environments are further divided into two folds, \textit{seen} and \textit{unseen}, to assess the generalization capacity.
For evaluation, we follow the standard evaluation protocol of the ALFRED benchmark \cite{shridhar2020alfred}.
The primary metric is the success rate, denoted by `SR,' which measures the percentage of completed tasks. %that the agent successfully completes.
Another metric is the goal-condition success rate, denoted by `GC,' which measures the percentage of satisfied goal conditions.
Finally, path-length-weighted (PLW) scores penalize SR and GC by the length of the actions that the agent takes.
We provide further details of this benchmark in the supplementary for space's sake.

% \vspace{-1em}
% \paragraph{Implementation Details.}
% % segmentation models, depth estimation models.
% % navigation algorithm.
% % hyperparameters set.
% We provide details of implementation in the supplementary.
% We will release all codes and models in a public repository.

\newcommand{\mcc}[2]{\multicolumn{#1}{c}{#2}}
\newcommand{\mcp}[2]{\multicolumn{#1}{c@{\hspace{30pt}}}{#2}}
\definecolor{Gray}{gray}{0.90}
\newcolumntype{a}{>{\columncolor{Gray}}r}
\newcolumntype{b}{>{\columncolor{Gray}}c}
\newcommand{\B}[1]{\textcolor{blue}{\textbf{#1}}}

\begin{table}[t!]
    \centering
    \resizebox{1.00\linewidth}{!}{
        \begin{tabular}{@{}lccaarr@{}}
            \toprule
            \multirow{2}{*}{\textbf{Model}} & \multirow{1}{*}{\textbf{Low}} & \multirow{1}{*}{\textbf{Tem.}} & \mcc{2}{\textbf{Test Seen}}  & \mcc{2}{\textbf{Test Unseen}} \\
                 & \multicolumn{1}{c}{\textbf{Inst.}} & \multicolumn{1}{c}{\textbf{Act.}} & \multicolumn{1}{b}{SR} & \multicolumn{1}{b}{GC} & \multicolumn{1}{c}{SR} & \multicolumn{1}{c}{GC} \\
            
            \cmidrule{1-7}

            % % \textbf{Templated Actions} \\[1pt]
            % \multicolumn{6}{l}{\textbf{Templated Actions}} \\
            
            % \cmidrule{1-8}

            % {\textbf{\method} (Ours)} \\ [1pt]
            % {~~~~ Templated actions}                    & $\B{46.44}$ ($19.14$) & $\B{57.93}$ ($24.56$) & $\B{46.50}$ ($\textbf{19.05}$) & $\B{59.00}$ ($\B{23.76}$) \\[1pt]
            % {\textbf{\method} w/ templates}         & $\B{-}$ ($-$) & $\B{-}$ ($-$) & $\B{-}$ ($\textbf{-}$) & $\B{-}$ ($\B{-}$) \\[1pt]
            
            {FILM \cite{min2021film}}               & \xmark & \cmark & $25.77$ ($10.39$)     & $36.15$ ($14.17$)     & $24.46$ ~~($9.67$)     & $34.75$ ($13.13$)     \\
            {Prompter \cite{inoue2022prompter}}     & \xmark & \cmark & $\textbf{49.38}$ ($\textbf{23.47}$) & $\textbf{55.90}$ ($\textbf{29.06}$) & $\underline{42.64}$ ($\underline{19.49}$)     & $\textbf{59.55}$ ($\textbf{25.00}$)     \\
            % {\textbf{\method} (Ours)}               & $\textbf{51.73}$ ($\textbf{22.34}$) & $\textbf{60.68}$ ($\textbf{26.91}$) & $\B{49.97}$ ($\B{21.52}$) & $\B{61.05}$ ($\textbf{25.27}$) \\[1pt]
            {\textbf{\method}$^*$}               & \xmark & \cmark & $\underline{46.64}$ ($\underline{20.81}$) & $\underline{55.29}$ ($\underline{25.47}$) & $\textbf{45.72}$ ($\textbf{20.15}$) & $\underline{57.25}$ ($\underline{24.73}$) \\ % + remember sliced
            
            % {\textbf{\method} (Ours)}               & $\textbf{53.03}$ ($\textbf{23.51}$) & $\textbf{61.26}$ ($\textbf{27.53}$) & $\B{49.05}$ ($\B{20.94}$) & $\B{60.45}$ ($\B{25.19}$) \\[1pt] % + remember sliced + erosion7
            
            \cmidrule{1-7}
            
            {FILM \cite{min2021film}}               & \cmark & \cmark & $28.83$ ($11.27$)     & $39.55$ ($15.59$)     & $27.80$ ($11.32$)     & $38.52$ ($15.13$)     \\
            {Prompter \cite{inoue2022prompter}}     & \cmark & \cmark & $\textbf{53.23}$ ($\textbf{25.81}$) & $\textbf{63.43}$ ($\textbf{30.72}$) & $\underline{45.72}$ ($\underline{20.76}$)     & $\underline{58.76}$ ($\textbf{26.22}$)     \\
            % {\textbf{\method} (Ours)}               & $\textbf{51.73}$ ($\textbf{22.34}$) & $\textbf{60.68}$ ($\textbf{26.91}$) & $\B{49.97}$ ($\B{21.52}$) & $\B{61.05}$ ($\textbf{25.27}$) \\[1pt]
            {\textbf{\method}$^*$}               & \cmark & \cmark & $\underline{50.62}$ ($\underline{22.61}$) & $\underline{59.40}$ ($\underline{27.49}$) & $\textbf{49.84}$ ($\textbf{22.61}$) & $\textbf{61.10}$ ($\textbf{27.00}$) \\ % + remember sliced
            
            % {\textbf{\method} (Ours)}               & $\textbf{53.03}$ ($\textbf{23.51}$) & $\textbf{61.26}$ ($\textbf{27.53}$) & $\B{49.05}$ ($\B{20.94}$) & $\B{60.45}$ ($\B{25.19}$) \\[1pt] % + remember sliced + erosion7
            
            \cmidrule{1-7}

            % {{\color{purple} \textbf{ECLAIR}}}               & \cmark & \cmark & ${52.58}$ (${23.09}$) & ${60.98}$ (${27.10}$) & ${50.36}$ (${21.59}$) & ${61.40}$ (${25.31}$) \\ % pred languages 
            
            {HLSM \cite{blukis2021persistent}}      & \xmark & \xmark & $25.11$ ~~($6.69$)    & $35.79$ ($11.53$)     & $16.29$ ~~($4.34$)    & $27.24$ ~~($8.45$)    \\
            % {AMSLAM \cite{jia2022learning}}         & $29.48$ ~~($3.28$)    & $40.88$ ~~($5.56$)    & $23.48$ ~~($2.36$)    & $34.64$ ~~($4.63$)    \\[1pt]
            % {LEBP \cite{liu2022lebp}}               & $25.53$ ~~($6.45$)    & $32.35$ ~~($9.99$)    & $24.26$ ~~($6.18$)    & $30.49$ ~~($9.89$)    \\[1pt]
            {LGS-RPA \cite{murray2022following}}    & \xmark & \xmark & $33.01$ ($\underline{16.65}$)     & $\underline{41.71}$ ($\textbf{24.49}$)     & $27.80$ ($\underline{12.92}$)     & $38.55$ ($\underline{20.01}$)     \\
            {EPA \cite{liu2022planning}}            & \xmark & \xmark & $\underline{39.96}$ ~~($2.56$)    & $44.14$ ~~($3.47$)    & $\underline{36.07}$ ~~($2.92$)    & $\underline{39.54}$ ~~($3.91$)    \\
            % {\textbf{\method} (Ours)}               & $\B{50.88}$ ($\textbf{20.16}$) & $\B{59.07}$ ($\textbf{24.23}$) & $\B{46.24}$ ($\textbf{18.66}$) & $\B{56.30}$ ($\textbf{22.58}$) \\[1pt] % pred languages (factored Meta Controller only)
            {\textbf{\method} (Ours)}         & \xmark & \xmark & $\textbf{47.36}$ ($\textbf{19.03}$) & $\textbf{54.38}$ ($\underline{23.78}$) & $\textbf{43.69}$ ($\textbf{17.64}$) & $\textbf{54.66}$ ($\textbf{22.76}$) \\ % pred languages (factored Meta Controller only) + remember sliced + not only put knife on sink 
            
            \cmidrule{1-7}
            
            {HLSM \cite{blukis2021persistent}}      & \cmark & \xmark & $29.94$ ~~($8.74$)    & $41.21$ ($14.58$)     & $20.27$ ~~($5.55$)    & $30.31$ ~~($9.99$)    \\
            {MAT \cite{ishikawa2022moment}}         & \cmark & \xmark & $33.01$ ~~($9.42$)    & $43.65$ ($14.68$)     & $21.84$ ~~($6.13$)    & $32.41$ ($10.59$)     \\
            {AMSLAM \cite{jia2022learning}}         & \cmark & \xmark & $29.48$ ~~($3.28$)    & $40.88$ ~~($5.56$)    & $23.48$ ~~($2.36$)    & $34.64$ ~~($4.63$)    \\
            % {LEBP \cite{liu2022lebp}}               & $25.53$ ~~($6.45$)    & $32.35$ ~~($9.99$)    & $24.26$ ~~($6.18$)    & $30.49$ ~~($9.89$)    \\[1pt]
            {LGS-RPA \cite{murray2022following}}    & \cmark & \xmark & $\underline{40.05}$ ($\underline{21.28}$)     & $\underline{48.66}$ ($\textbf{28.97}$)     & $\underline{35.41}$ ($\textbf{22.76}$)     & $\underline{45.24}$ ($\underline{22.76}$)     \\
            % {EPA \cite{liu2022planning}}            & $39.96$ ~~($2.56$)    & $44.14$ ~~($3.47$)    & $\textbf{36.07}$ ~~($2.92$)    & $39.54$ ~~($3.91$)    \\[1pt]
            % {\textbf{\method} (Ours)}               & $\B{50.88}$ ($\textbf{20.16}$) & $\B{59.07}$ ($\textbf{24.23}$) & $\B{46.24}$ ($\textbf{18.66}$) & $\B{56.30}$ ($\textbf{22.58}$) \\[1pt] % pred languages (factored Meta Controller only)
            % {\textbf{\method} (Ours)}               & \cmark & \xmark & $\textbf{51.92}$ ($\underline{21.53}$) & $\textbf{60.18}$ ($\underline{25.71}$) & $\textbf{45.85}$ ($\underline{18.94}$) & $\textbf{56.49}$ ($\textbf{23.68}$) \\ % pred languages (factored Meta Controller only) + remember sliced

            {\textbf{\method} (Ours)}               & \cmark & \xmark & $\textbf{51.79}$ ($\textbf{21.60}$) & $\textbf{60.50}$ ($\underline{25.88}$) & $\textbf{46.11}$ ($\underline{19.45}$) & $\textbf{57.33}$ ($\textbf{24.06}$) \\ % pred languages 
            
            \cmidrule{1-7}
            
            {Human}                                 & & & \multicolumn{1}{b}{-} & \multicolumn{1}{b}{-} & $91.00$ ($85.80$) & $94.50$ ($87.60$) \\
            \bottomrule
        \end{tabular}
    }
    \vspace{-0.5em}
    \caption{
        \textbf{Comparison with the state of the arts.}
        The path-length-weighted (PLW) metrics are given in the parentheses for each value.
        The highest and second-highest values per fold and metric are shown in \textbf{bold} and \underline{underline}, respectively.
        `Low Inst.' refers to the step-by-step instructions aligned to the respective sub-goals.
        `Tem. Act.' refers to `templated action' sequences designed in \cite{min2021film}.
        {\cmark/\xmark} denotes the corresponding module is used/not used, respectively.        
        `CAPEAM$^*$' denotes our agents using the templated actions without action planning by our learned model.
        % {\color{purple} \textbf{ECLAIR}}, a slight variant of \textbf{\method}$^*$ including hyperparameter tuning, denotes our submission for the 1$^{st}$ generalist language grounding agents challenge in CVPRW'23.
    }
    % \vspace{-1em}
    \label{tab:comparison_with_sota}
\end{table}

\begin{table}[t!]
    \centering
    \resizebox{\columnwidth}{!}{
        \begin{small}
            \begin{tabular}{@{}cccbbcc@{}}
                \toprule
                \multirow{2}{*}{~~~\#}
                    & \multirow{2}{*}{CAP} & \multirow{2}{*}{EAM}
                    & \mcc{2}{\cellcolor[HTML]{FFFFFF} \textbf{Test Seen}} &  \mcc{2}{\textbf{Test Unseen}} \\
                    & & & \multicolumn{1}{b}{SR} & \multicolumn{1}{b}{GC} & \multicolumn{1}{c}{SR} & \multicolumn{1}{c}{GC} \\
                \midrule
                \multicolumn{1}{c}{($a$)} & \cmark & \cmark & $51.79$ ($21.60$) & $60.50$ ($25.88$) & $46.11$ ($19.45$) & $57.33$ ($24.06$) \\
                \midrule
                \multicolumn{1}{c}{($b$)} & \xmark & \cmark & $49.45$ ($20.77$) & $58.94$ ($25.43$) & $42.25$ ($17.79$) & $53.93$ ($22.30$) \\
                \multicolumn{1}{c}{($c$)} & \cmark & \xmark & $46.44$ ($18.12$) & $56.37$ ($23.24$) & $41.66$ ($16.03$) & $52.88$ ($21.40$) \\
                \multicolumn{1}{c}{($d$)} & \xmark & \xmark & $45.34$ ($17.55$) & $56.62$ ($22.99$) & $39.83$ ($15.40$) & $51.97$ ($20.15$) \\
                \bottomrule
            \end{tabular}
        \end{small}
    }
    \vspace{-0.5em}
    \caption{
        \textbf{Ablation study for each proposed component.} % of our \method.}
        CAP denotes `Context-Aware Planning.'
        EAM denotes `Environmental-aware Memory.'
        The ablation of either or both CAP and EAM leads to noticeable performance drops.
    }
    \vspace{-1em}
    \label{tab:ablation_proposed_components}
\end{table}

\subsection{Comparison with the State of the Art}
\label{subsec:comparison_with_sota}
We present a quantitative analysis of our method and prior arts in Table \ref{tab:comparison_with_sota}.
For a fair comparison, we compare our method with prior arts that incorporate semantic spatial representation constructed followed by depth estimation.
Following the recent approaches \cite{blukis2021persistent,min2021film}, we compare methods that use only a high-level goal statement (\ie, without low-level instructions, denoted by `Low Inst.' by {\xmark}).
In addition, we compare the models that generate action sequences using prior knowledge of tasks and environments with the `action template' (`Tem. Act.' by {\cmark}) \cite{min2021film,inoue2022prompter}.

We first investigate the performance when the hand-designed action sequence templates are combined with our agent ({\cmark} in `Tem. Act.'), which is an ablated version of our model.
We observe that our agent outperforms all prior arts in novel environments in terms of success rates, which is the main metric of the benchmark.
We observe \cite{inoue2022prompter} yields better performance in seen environments compared to our agent.
We believe that this might be attributed to the strategies to enhance spatial perception such as pose adjustment based on accurate perception models (\eg, depth estimators, \etc) that generally perform well in seen environments.
Nevertheless, we observe that agent has less performance gap between seen and unseen environments, implying better generalization of our agent to unseen environments.

We then investigate the performance without using the templated action sequences ({\xmark} in `Tem. Act.'). 
We observe that our method outperforms all prior arts by large margins in SR and GC for both seen and unseen environments.
As we consistently observe the improvements with and without the low-level instructions, this would imply that our method does not heavily rely on the detailed description of a task.
Note that \cite{murray2022following} collects an additional dataset with the human-in-the-loop process for interaction failure recovery.
With the additional expensive supervision, we observe that \cite{murray2022following} achieves better PLW scores by efficiently completing tasks than \method, and the comparison is not quite fair.

\subsection{Ablation Study}
To investigate the benefits of each proposed component of our \method, we conduct a quantitative ablation study and summarize the result in Table \ref{tab:ablation_proposed_components}.

\begin{figure*}[t!]
    \centering
    \includegraphics[width=0.48\linewidth]{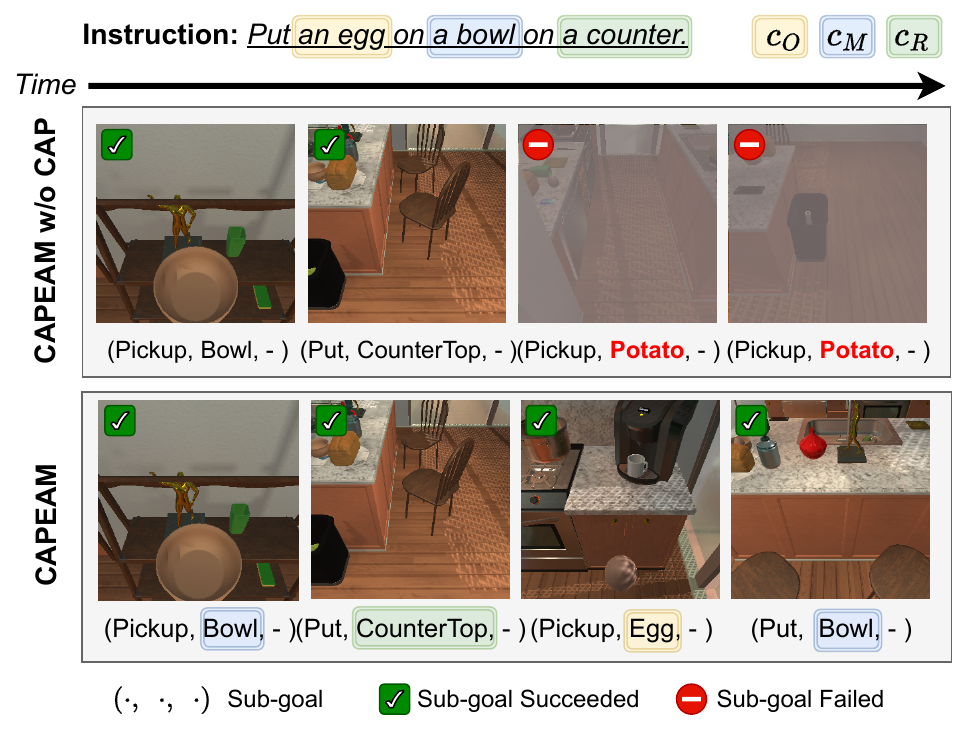}
    \hspace{1em}
    \includegraphics[width=0.48\linewidth]{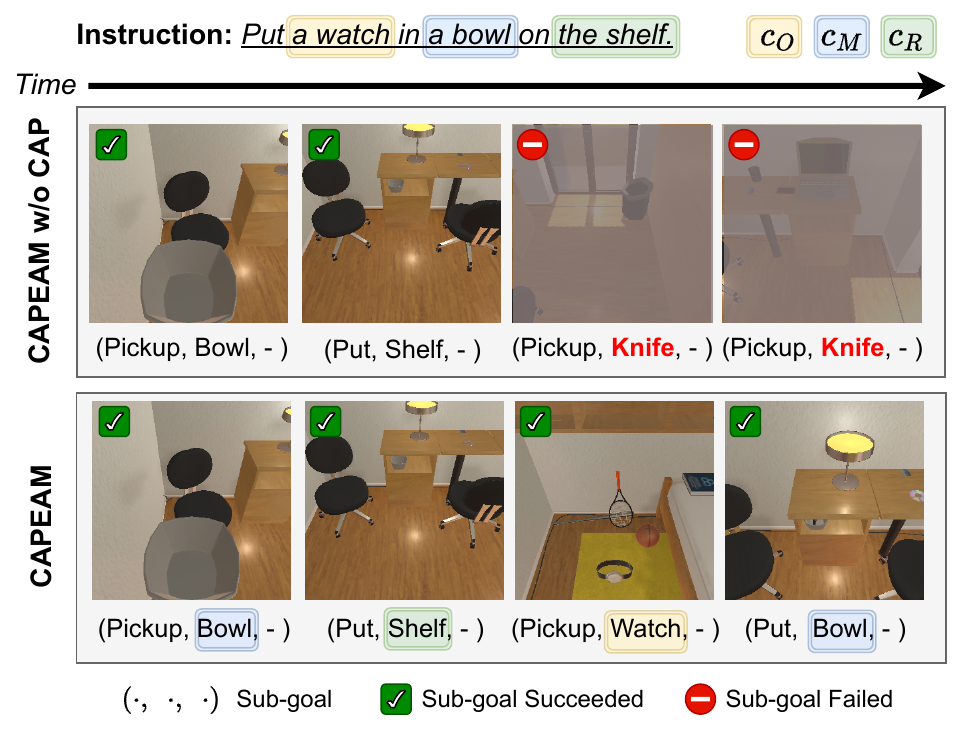}
    \vspace{-.5em}
    \caption{
        \textbf{Benefit of Context-Aware Planning (CAP).}
        In the two qualitative examples, the `contexts' are denoted by $c_O$ in yellow, $c_M$ in blue, and $c_R$ in green colored boxes.
        While our \method plans a sub-goal sequence with task-relevant objects, `\method w/o CAP' interacts with task-irrelevant objects (\ie, {\color{red} Potato} or {\color{red} Knife}) and consequently fails. % at the task.
    }
    \vspace{-1em}
    \label{fig:qualitative_cap}
\end{figure*}

\begin{table}[t!]
    \centering
    \resizebox{0.90\columnwidth}{!}{
        \begin{small}
            \begin{tabular}{@{}cccbcc@{}}
                \toprule
                \multirow{1}{*}{~~~\#}
                    & \multirow{1}{*}{CAP} & \multirow{1}{*}{Param\#} & \mcc{1}{\textbf{Valid. Seen}} & \mcc{1}{\textbf{Valid. Unseen}} \\
                \midrule
                \multicolumn{1}{c}{($a$)} & \cmark & $712.60$M & $83.05$ & $77.34$ \\
                \midrule
                \multicolumn{1}{c}{($b$)} & \xmark & $651.01$M & $79.27$ & $74.91$ \\
                \multicolumn{1}{c}{($c$)} & \xmark & $164.95$M & $80.12$ & $74.18$ \\
                \bottomrule
            \end{tabular}
        \end{small}
    }
    \vspace{-0.5em}
    \caption{
        \textbf{Accuracy of the predicted action sequence compared to the ground-truth by the CAP module \vs na\"ive model size increase.}
        CAP refers to the `Context-Aware Planning.'
        Param\# denotes the number of parameters to be learned for planning.
        (b) and (c) share the same LSTM-based architecture of (a) but not use context (\ie, a set of task-relevant objects).
        CAP noticeably improves the success rate ((a) $\rightarrow$ (c)) but this is not simply from the model size increase ((c) $\rightarrow$ (b)).
    }
    \vspace{-1em}
    \label{tab:ablation_capacity}
\end{table}

\vspace{-1em}
\paragraph{Without Context-Aware Planning.}
First, we ablate the CAP from our method and the agent therefore learns a monolithic policy that directly maps natural language instructions to a sequence of agent-executable actions.
At times, the agent attempts to interact with inappropriate objects to tasks.
The result of the task may differ from the intended goal, leading to task failure.
This eventually leads to noticeable performance drops ($-2.34\%$, $-3.86\%$ in SR) in both seen and unseen splits as evidenced in (\#(a) \textit{vs.} \#(b)).

Some may argue that the performance gain by the CAP may come from model size increase as we use separate networks for context prediction and sub-goal frame sequence generation. %  to generate a sequence of sub-goals, resulting in the increased capacity size.
For this, we learn the same policy but with more learnable parameters such as the size of hidden states and embeddings to closely approach to the size of the final model and summarize the result in Table \ref{tab:ablation_capacity}.
We observe that even with the similar number of parameters ($651.01$M compared to $712.60$M), the planning accuracy of the policy with more parameters remains similar or even slightly drops in the seen split from the one with the smaller network ($164.95$M). 
It implies that the performance gain by the CAP may not be attributed to simple model size increase.

\vspace{-0.5em}
\paragraph{Without Environment-Aware Memory.}
We then ablate the EAM from our agent.
% In this case, the agent has to conduct its actions solely based on the current state as the agent does not preserve any information about the changed states of objects and their masks.
Without EAM, the agent has to conduct its actions based on the current state as the agent preserves only limited information about the changed states of objects and their masks.
Due to the lack of environmental information, the agent may perform undesired actions (\eg, move an already relocated object) and thus fail. % at the task.
We observe significant performance drops ($-5.35\%$ and $-4.45\%$ in SR) in both seen and unseen environments (\#(a) \textit{vs.} \#(c)).

\vspace{-0.5em}
\paragraph{Without Both.}
Without any of the proposed components, the agent may interact with irrelevant objects with limited past environmental information. %, the agent has more chances to fail at the task due to undesired task results than the agents equipped with either or both CP and EM.
As expected, our agent without CAP and EAM achieves the lowest performance among the agents equipped with either or both (\#(d) \textit{vs.} \#(a, b, c)).
Moreover, we observe that using both CAP and EAM improves our agent more than using either of them ((\#(d) $\rightarrow$ \#(b, c)) \textit{vs.} (\#(d) $\rightarrow$ \#(a))).
This implies that the CAP and the EAM are complementary to each other.

\subsection{Qualitative Analysis}

\paragraph{Context-Aware Planning.}
To illustrate the benefit of context-aware planning (CAP), we present two qualitative examples in Figure ~\ref{fig:qualitative_cap}.
The left example shows that a sub-goal planner without the CAP may generate sub-goals including irrelevant objects (\ie, {\color{red} Potato}), even when it is not mentioned in the given instruction.
The agent continues searching for a knife for the limited number of steps and fails.
But when we use the context predictor, the sub-goal planner correctly infer the task-relevant objects and constructs the sequence with them (Sec.~\ref{subsection:context_aware_multi_level_planning}); the context predictor outputs task-relevant objects from the given human-described instruction such as an `egg,' denoted as $o_O$, a `bowl', the container to hold the object as $o_M$, and a `counter', the place to put the object as $o_R$.
The task-relevant objects that is correctly inferred by our model, the sub-goal planner can generate a desirable sub-goal sequence and leads to successful task completion.

In addition, the right example in the Figure \ref{fig:qualitative_cap} shows that our agent without CAP may predict irrelevant objects (\ie, Knife) to the task (\ie, ``Put a watch in a bowl on the shelf.'').
As the agent without CAP, denoted by `\method w/o CAP,' tries to find the unintended object (\ie, Knife) which is not present in this room, the agent continues to explore the environment and eventually fails to reach the target object.
In contrast, our agent with CAP, denoted by `\method,' can generate a sequence of executable actions with relevant objects.
By conducting all the actions with the intended objects, the agent finally succeeds in the task. %, implying the importance of CAP.

\vspace{-1em}
\paragraph{Environment-Aware Memory.}

\begin{figure}[t!]
    \centering
    \includegraphics[width=\columnwidth]{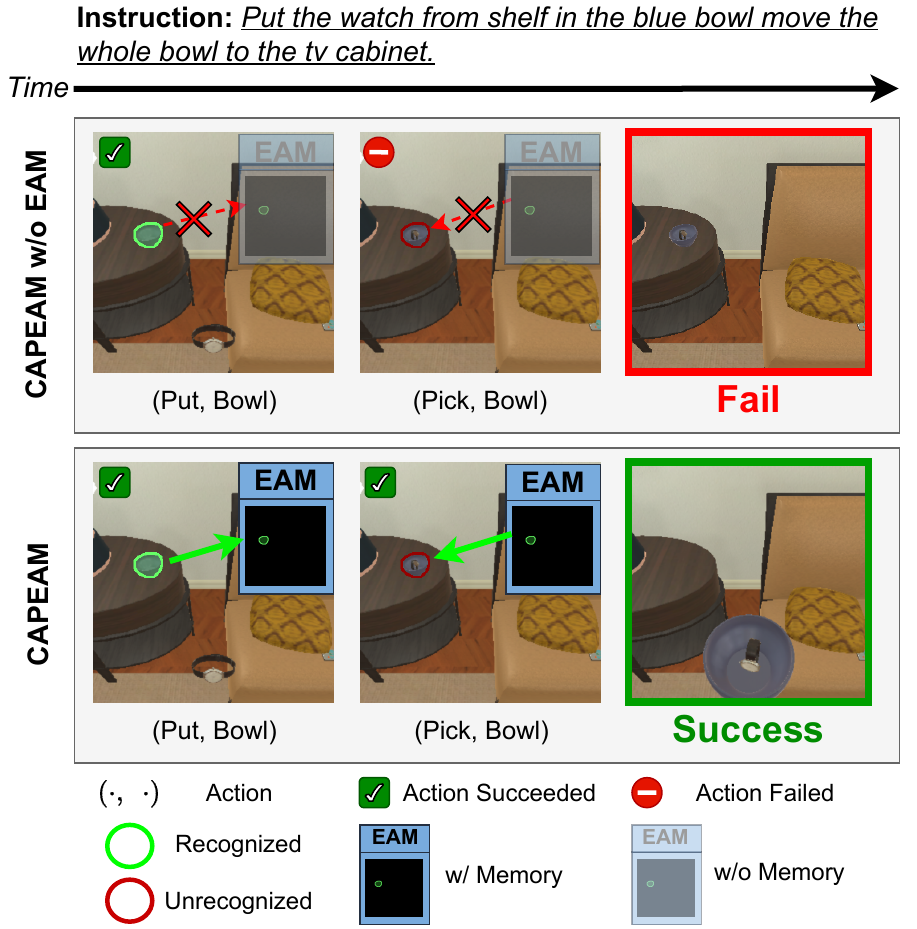}
    \vspace{-1.5em}
    \caption{
        \textbf{Benefit of the `Retrospective Object Recognition' in the Environment-Aware Memory (EAM).}
        {\color{green}$\rightarrow$} indicates the agent can preserve the object masks in EAM and utilize them while {\xmark} cannot.
        % While our agent 
        `\method w/o EAM' fails in interaction since it cannot recognize the bowl.
        In contrast, `\method' can exploit the preserved bowl's mask and therefore succeeds in the task.}
    \label{fig:qualitative_em}
    \vspace{-1em}
\end{figure}

We now conduct a qualitative analysis to assess the impact of retrospective object recognition (Sec.~\ref{subsec:Retrospective_Object_Recognition}). 
It allows the agent to continue interacting by utilizing a previously saved mask even when it cannot recognize the object.

First, we investigate the benefit of the 'Retrospective Object Recognition' module in the EAM.
Prior arts \cite{suvaansh2023multi, min2021film} may miss interactions with unrecognized objects because they have limited access to past information and typically rely on current information during the interaction.
For instance, in Figure \ref{fig:qualitative_em}, the memoryless model is incapable of detecting the bowl's mask because of occlusion caused by the watch placed on top of it when it tries to pick up the bowl.
Thus, they may initially encounter challenges with interaction, given their incapability to obtain a mask for interaction.
On the contrary, \method with the memory stores the bowl's mask from the previous action and uses it for interaction.
Consequently, even when the agent cannot perceive the bowl, it can interact with it by the saved mask.

\begin{figure}[t!]
    \centering
    \includegraphics[width=\columnwidth]{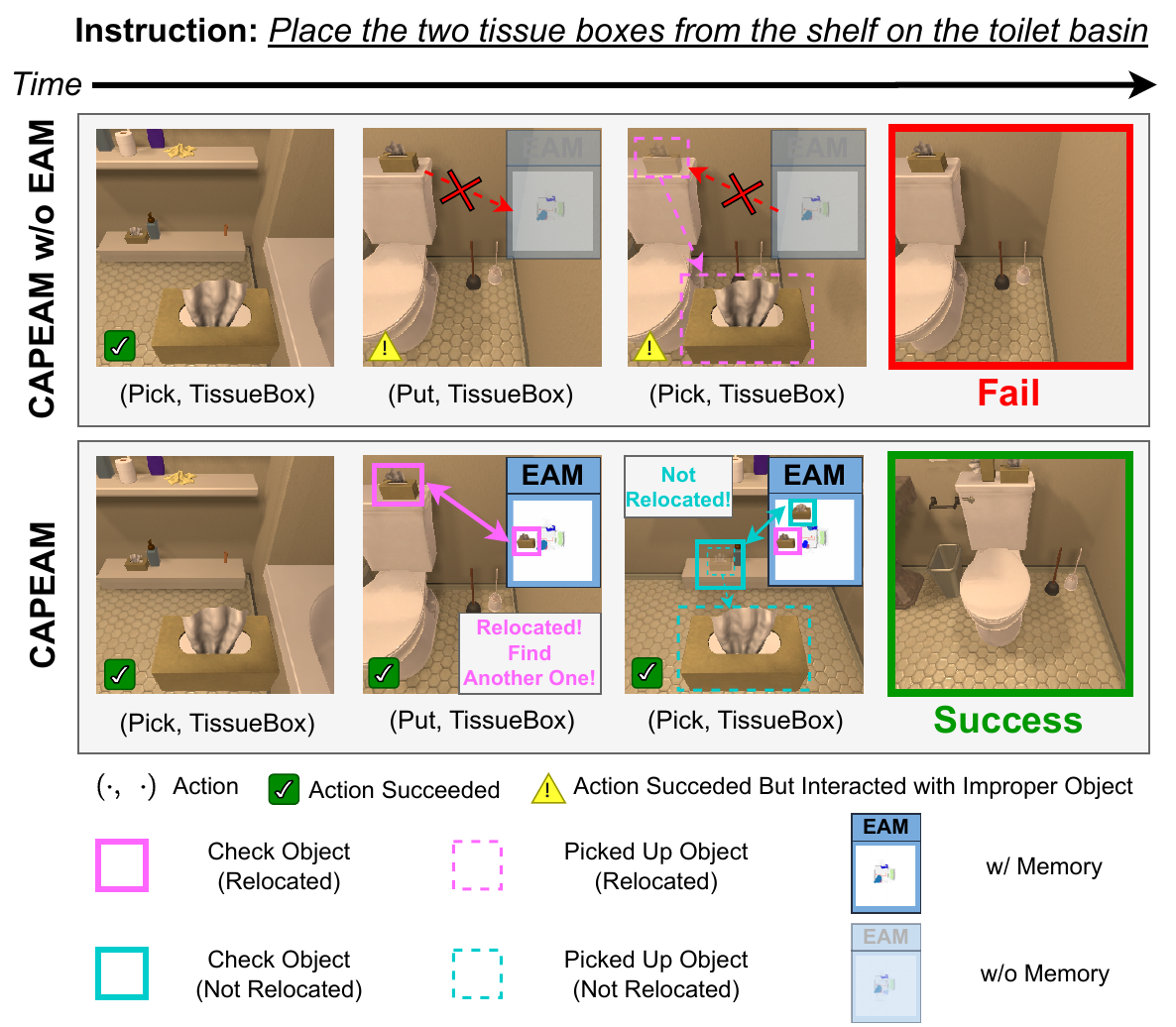}
    \vspace{-1.5em}
    \caption{
        \textbf{Benefit of the `Object Relocation Tracking' in the EAM.}
        While our agent without EAM (`\method w/o EAM') interacts with the already relocated object (`TissueBox') as it does not keep track of the relocated location (denoted by the dashed {\color{red} $\rightarrow$} and {\color{red} \xmark}), EAM (`\method') allows to avoid interacting with the already relocated one and therefore interact with the intended one.}
    \label{fig:qualitative_em_ort}
    \vspace{-1em}
\end{figure}

Finally, we investigate the benefit of the 'Object Relocation Tracking' module in the EAM. 
Figure \ref{fig:qualitative_em_ort} illustrates its benefit of keeping track of the relocated objects and preventing re-relocation of the same objects (Sec. \ref{subsection:environment_aware_memory}). %{\color{blue} 3.2}).
As observed in `\method' in the figure, after relocating a target object (\ie, `TissueBox'), the agent remembers the relocated object's location and avoids interacting with that already relocated tissue box.
On the contrary, as observed in `\method w/o EAM,' our agent without EAM does not keep track of the relocated object's location.
Thus, it interacts with the already relocated object again and eventually, this leads to task failure.

\section{Conclusion}

We propose \method that incorporates \emph{context} in planning and to remember environmental changes in memory for embodied AI agents. 
It improves navigation and object interaction by avoiding unnecessary exploration and correct planning of actions with appropriate objects to interact with.
We empirically validate the benefit of the proposed modules in ALFRED by showing that the proposed method outperforms existing methods, especially in unseen environments even without human-designed plan templates.

\vspace{-1em}
\paragraph{Limitation and Future Work.}
Our context-aware planning fixes the anticipated context during a task execution as an inductive bias.
But the context may change even in a single task execution. 
A prospective avenue for future investigation lies in the modification of context in response to input from environments, thereby enhancing adaptability.
% Looking ahead, we suggest exploring planning with context equipped with memory as a promising avenue for future research in this exciting field.

\vspace{-1.5em}
{\footnotesize
\begin{singlespace}
\paragraph{\footnotesize Acknowledgment.} This work is partly supported by the NRF grant (No.2022R1A2C4002300) 20\%, IITP grants (No.2020-0-01361, AI GS Program (Yonsei University) 5\%, No.2021-0-02068, AI Innovation Hub 5\%, 2022-0-00077 10\%, 2022-0-00113 10\%, 2022-0-00959 10\%, 2022-0-00871 20\%, 2022-0-00951 20\%) funded by the Korea government (MSIT).
\end{singlespace}
}

{\small
\bibliographystyle{ieee_fullname}
\bibliography{egbib}
}

\clearpage
\appendix

\definecolor{Gray}{gray}{0.90}
\newcolumntype{a}{>{\columncolor{Gray}}r}
\newcolumntype{b}{>{\columncolor{Gray}}c}

\newcommand{\bmp}[1]{\textcolor{blue}{#1}} % blue main paper
\newcommand{\orange}[1]{\textcolor{orange}{#1}} % blue main paper

\noindent\textbf{Note:} {\color{blue}Blue} characters denote the main paper's reference.

\section{Details of ALFRED Benchmark}
The task goal is to generate a sequence of actions and object masks for interaction with the corresponding objects such that an agent satisfies all conditions that define the task to be successful.
If the agent does not satisfy even a single condition, the agent is considered failed at the task.

To train and assess such agents, the benchmark consists of three splits; `train,' `validation,' and `test.'
Agents can be trained with the `train' split and validate their approaches in the `validation' split with the ground-truth information of the tasks in those splits.
The agents are then evaluated in the `validation' and `test' split but they do not have any access to the ground-truth information of the tasks.

To complete a task, the agent receives a goal statement that provides a high-level description of the task's goal and step-by-step instructions that provide detailed explanations for how to complete respective steps (\ie, sub-goals).
Given the goal statement and step-by-step instruction, the agent receives an egocentric RGB image in the shape of $300\times300$ and takes an action and an object mask for each time step.

% output space (5 nav actions, 7 int actions, 1 stop action)
The action space of the agent consists of five navigation actions, seven interaction actions, and a \textsc{Stop} action that indicates the termination of task completion.
The navigation actions are \textsc{MoveAhead} for moving ahead, \textsc{RotateRight}/\textsc{RotateLeft} for rotating right/left to 90$^{\circ}$, and \textsc{LookUp}/\textsc{LookDown} for looking up/down to 15$^{\circ}$.
The interaction actions are \textsc{PickupObject} for picking up an object (\eg, an apple, a potato, \etc), \textsc{PutObject} for putting an object in a receptacle (\eg, a countertop, a desk, \etc), \textsc{OpenObject}/\textsc{CloseObject} for opening/closing an object (\eg, a cabinet, a drawer, \etc), \textsc{ToggleObjectOn}/\textsc{ToggleObjectOff} for turning on/off an object (\eg, a microwave, a lamp, \etc), and \textsc{SliceObject} for slicing an object (\eg, a tomato, a bread, \etc).
For object interaction, the agent has to additionally predict a binary object mask in the same shape as the RGB image (\ie, $300 \times 300$), and the agent interacts with the object to which the highest number of the mask pixels belongs.

For evaluation, the benchmark uses three types of metrics; success rate (SR), goal-condition success rate (GC), and PLW scores (PLWSR and PLWGC).
The primary metric is the success rate (SR) which measures the percentage of completed tasks.
This metric indicates the task completion ability of the agent.
Another metric is the goal-condition success rate (GC) which measures the percentage of satisfied goal conditions.
This metric indicates the partial task completion ability of the agent.
Finally, path-length-weighted (PLW) scores penalize SR and GC by the length of the agent's actions.
This metric indicates the ability to complete tasks efficiently.
% We provide further details of this benchmark in the supplementary for space's sake.

\section{Additional Qualitative Analysis}
% We further provide additional qualitative analyses for Context-aware Planning (CP) and Environment-aware Memory (EM) in Figure \ref{fig:qualitative_cap}-\ref{fig:qualitative_em_olc}.

\subsection{Context-Aware Planning}
\begin{figure}[t!]
    \centering
    \includegraphics[width=.95\columnwidth]{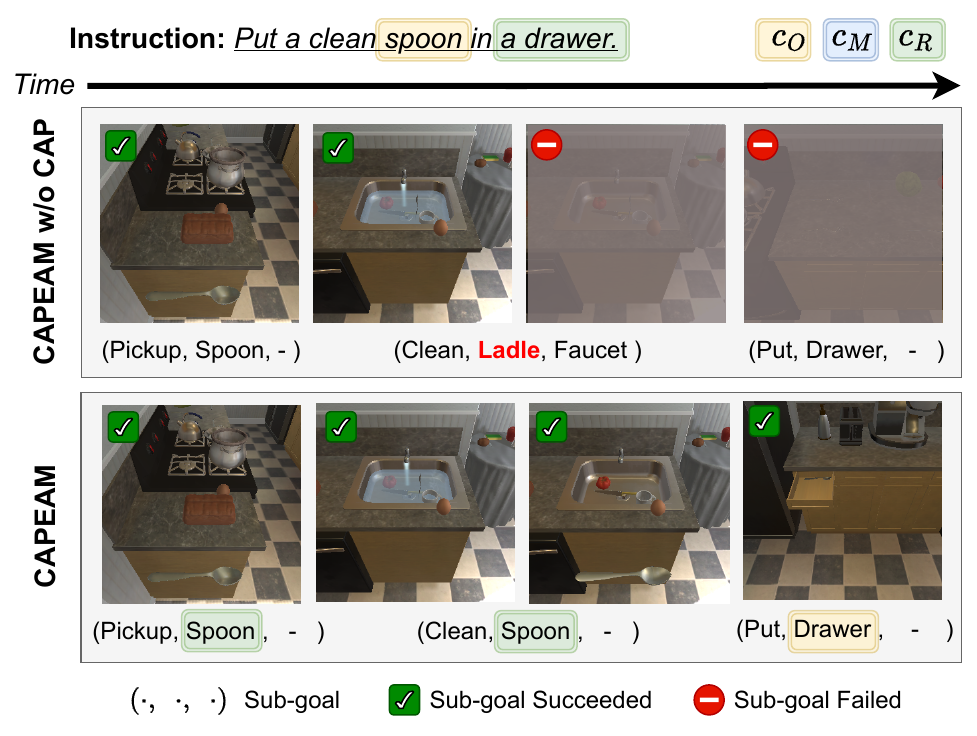}
    \vspace{-1em}
    \caption{
        \textbf{Another qualitative example of our agent with and without the `Context-aware Planning' (CAP).}
        The elements of the `context' are denoted by $c_O$ in yellow, $c_M$ in blue, and $c_R$ in green.
        Our method (\method) plans a sequence of sub-goals with task-relevant objects.
        However, `\method w/o CAP' interacts with task-irrelevant objects (\ie, {\color{red} Ladle}), leading to task failure.
    }
    \vspace{-1em}
    \label{fig:qualitative_cap_supp2}
\end{figure}

In the same manner as in Figure {\color{blue} 5} and {\color{blue} 6} in the main paper, we provide another qualitative example in Figure \ref{fig:qualitative_cap_supp2}.
Figure \ref{fig:qualitative_cap_supp2} also shows similar results for the task (``Put a clean spoon in a drawer.'').
`\method w/o CAP' can pick up the relevant object, a spoon, and clean it as described in the instruction but tries to pick up a task-irrelevant object, a ladle, instead of the spoon due to the wrong object prediction, leading to task failure.
On the other hand, CAP enables our agent to keep interacting with the task-relevant object (\ie, the spoon) and the agent finally puts the spoon in a drawer as described in the instruction, implying the impact of CAP.

\begin{figure}[t!]
    \centering
    \includegraphics[width=\columnwidth]{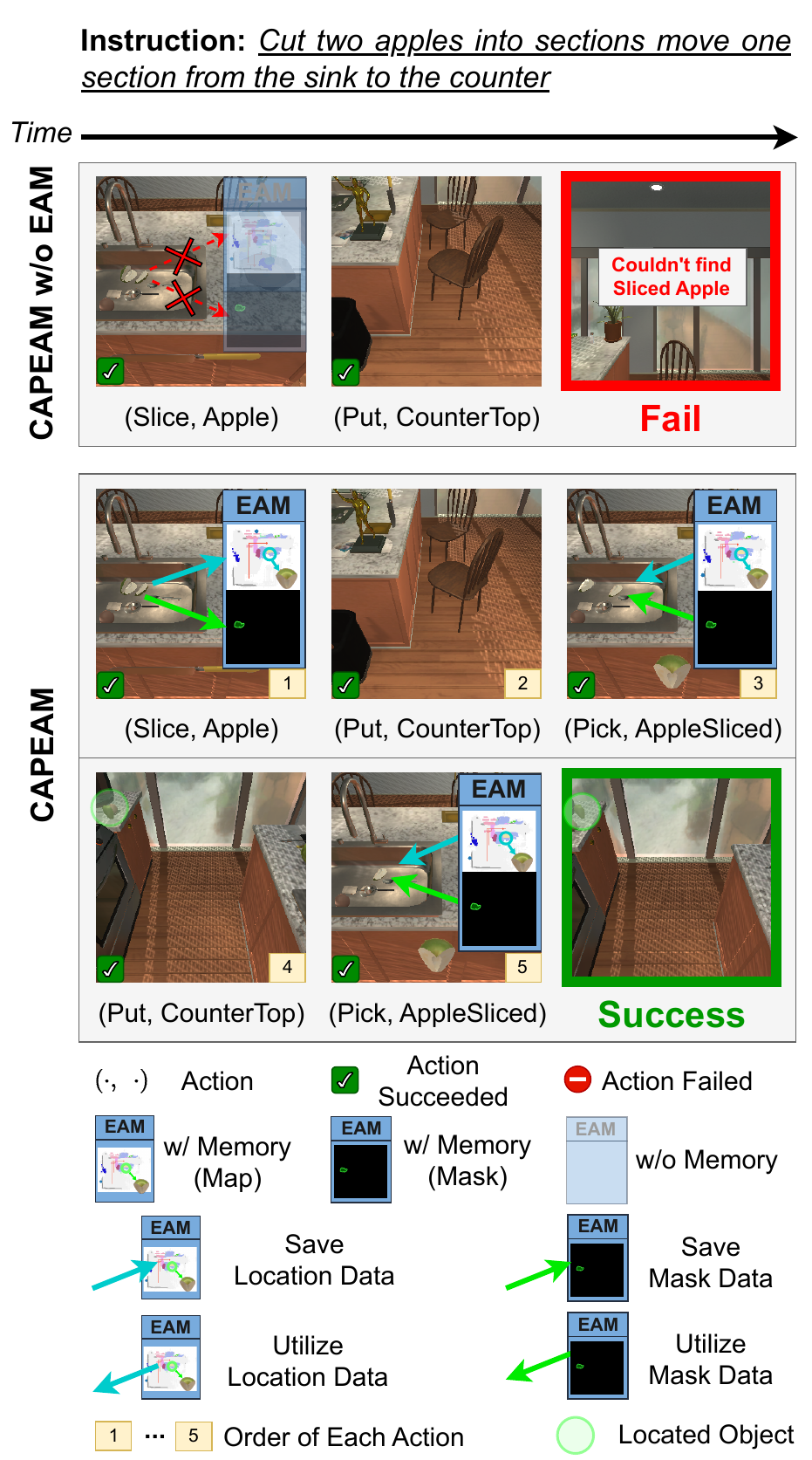}
    \vspace{-1.5em}
    \caption{
        \textbf{Another qualitative example of the benefit of `Environment-aware Memory' (EAM) (for `Object Location Caching').}
        While our agent without EAM (`\method w/o EM') succeeds in slicing the target object (\ie, `Apple') and putting the knife on the countertop, the agent does not remember the location of the sliced object and therefore it has to explore the environment again to reach the sliced object, eventually leading to task failure.
        However, after slicing the apple and putting the knife on the countertop, our agent with EAM (`\method') remembers the location of the object with the changed state (\ie, a `sliced' apple) with its mask and thus our agent can navigate back to its location and successfully move two sliced apples to the designated countertop.
    }
    \label{fig:qualitative_em_olc}
    \vspace{-1em}
\end{figure}

\subsection{Environment-Aware Memory}
Like Figure {\color{blue} 7} and {\color{blue} 8} in the main paper, Figure \ref{fig:qualitative_em_olc} shows the impact of Object Location Caching (Sec. {\color{blue} 3.2}).
This enables our agent to preserve the location and mask of an object with a changed state so that the agent can utilize the information when necessary.
Although our agent without EAM (`\method w/o EAM') successfully slices an object (here, an apple) and puts the knife back on the countertop, the agent does not memorize the object's location and therefore it has to explore the environment to navigate to the apple whose state changes (\ie, sliced).
In this example, the agent fails at reaching the sliced object and eventually fails at the task.
However, our agent equipped with EAM (`\method') preserves the location and mask of the sliced object and can navigate back to the saved location, which reduces unnecessary exploration and possible task failure.
In this example, the agent successfully reaches the sliced objects and moves them to the designated countertop.

\section{A Video with Additional Qualitative Results}
We also provide an additional qualitative analysis in the attached video files. % (\textit{video.mp4}).
The video contains two qualitative examples of the ablation of each `Context-Aware Planning (CAP)' and `Environment-Aware Memory (EAM),' respectively.
The agents for the video take as input only the goal statement (\ie, no step-by-step instructions).

\subsection{Example 1: Context-Aware Planning}
For the task ``Throw two bars of soap in the trash bin,'' our agent without CAP (\textit{noCAP.mp4}) predicts an object irrelevant to the task (\ie, SoapBottle), which is out of context (\ie, $c_O=\text{SoapBar}$, $c_M=\text{None}$, and $c_R=\text{GarbageCan}$).
Although the agent succeeds in implementing the agent-executable actions, they lead to an undesired result (\ie, move a soap bottle, not a soap bar) and thus the agent fails.

However, our agent equipped with CAP (\textit{CAP.mp4}) correctly plans to pick two `SoapBar' objects and succeeds in taking all the planned executable actions.
As the actions lead to the desired result (\ie, move two soap bars), the agent finally succeeds, implying the CAP's efficacy.

\subsection{Example 2: Environment-Aware Memory}
For the task ``Put a cup with a fork in it in the sink,'' while our agent without EAM (\textit{noEAM.mp4}) can predict a cup without a fork, it cannot recognize the cup with a different visual appearance (\ie, the cup with the fork) and thus the agent interacts with the wrong object, ``Pan,'' and fails. % leading to task failure.

On the other hand, after putting the fork in the cup, our agent with EAM (\textit{EAM.mp4}) can still recognize the cup thanks to the preserved mask in EM used as an approximation of the mask with a different visual appearance.
The agent succeeds in conducting all predicted actions and finally completes the task, which implies the impact of EAM.

\end{document}